%% file: main.tex
\documentclass[letterpaper,10pt,conference]{ieeeconf}  

\IEEEoverridecommandlockouts                              

\usepackage{mymacros}
\usepackage{times} 
\usepackage{graphicx}

\usepackage[T1]{fontenc}
\usepackage{color}
\usepackage{hyperref}
\usepackage{pifont}

\let\labelindent\relax
\usepackage{enumitem}

\title{\LARGE \bf
Switching Head--Tail Funnel UNITER for\\
Dual Referring Expression Comprehension with Fetch-and-Carry Tasks
}

\author{Ryosuke Korekata$^{1}$, Motonari Kambara$^{1}$, Yu Yoshida$^{1}$, Shintaro Ishikawa$^{1}$, Yosuke Kawasaki$^{1}$,\\
Masaki Takahashi$^{1}$, and Komei Sugiura$^{1}$
\thanks{
    \small $^{1}$The authors are with Keio University, 3-14-1 Hiyoshi, Kohoku, Yokohama, Kanagawa 223-8522, Japan.
    {\tt\small rkorekata@keio.jp}
}
}

\begin{document}

\maketitle
\thispagestyle{empty}
\pagestyle{empty}

\input{abstract}
\input{section1}
\input{section2}
\input{section3}
\input{section4}
\input{section5}
\input{section6}
\input{section7}


\vspace{-0.853mm}
\section*{ACKNOWLEDGMENT}
\vspace{-1mm}
This work was partially supported by JSPS KAKENHI Grant Number 20H04269, JST CREST, and NEDO.
\vspace{-0.5mm}
\bibliographystyle{IEEEtran}
\bibliography{reference}

\end{document}

%% file: abstract.tex
\begin{abstract}
This paper describes a domestic service robot (DSR) that fetches everyday objects and carries them to specified destinations according to free-form natural language instructions.
Given an instruction such as ``Move the bottle on the left side of the plate to the empty chair,'' the DSR is expected to identify the bottle and the chair from multiple candidates in the environment and carry the target object to the destination.
Most of the existing multimodal language understanding methods are impractical in terms of computational complexity because they require inferences for all combinations of target object candidates and destination candidates.
We propose Switching Head--Tail Funnel UNITER, which solves the task by predicting the target object and the destination individually using a single model.
Our method is validated on a newly-built dataset consisting of object manipulation instructions and semi photo-realistic images captured in a standard Embodied AI simulator.
The results show that our method outperforms the baseline method in terms of language comprehension accuracy.
Furthermore, we conduct physical experiments in which a DSR delivers standardized everyday objects in a standardized domestic environment as requested by instructions with referring expressions.
The experimental results show that the object grasping and placing actions are achieved with success rates of more than 90\%.
\end{abstract}

%% file: section1.tex
\section{Introduction
\label{intro}
}
\vspace{-1mm}

\begin{figure}[t]
    \centering
    \includegraphics[height=8.8cm]{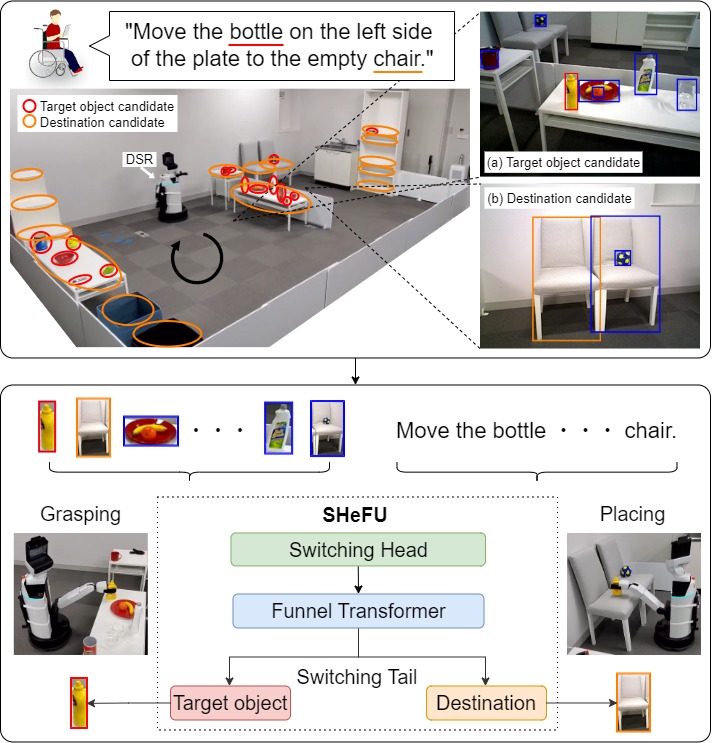}
    \caption{\small Overview of SHeFU. The red, orange, and blue bounding boxes in the images (a) and (b) represent the target object, the destination, and surrounding objects or furniture pieces, respectively.}
    \label{fig:eye_catch}
    \vspace{-4mm}
\end{figure}

In today’s aging society, there is a growing demand for assistance in everyday life.
This has led to a scarcity of home caregivers, which is becoming a social issue.
As a solution, domestic service robots (DSRs) that can physically support care recipients are attracting considerable attention~\cite{yamamoto2019development}.
However, the ability of the DSRs to comprehend natural language instructions still requires improvement.

In this study, we aim to develop a DSR that fetches everyday objects and carries them to specified locations according to free-form natural language instructions.
Fig.~\ref{fig:eye_catch} shows an overview of our method.
Specifically, for an instruction such as ``Move the bottle on the left side of the plate to the empty chair,'' the DSR should identify the bottle as the target object and the chair as the destination from among the surrounding objects or furniture pieces.
The DSR should then grasp the bottle and place it on the chair.

The instructions given by humans are often ambiguous, making it challenging for DSRs to identify the target object and the destination.
In fact, \cite{shridhar2020alfred} reported that humans scored 91.0\% on the ALFRED dataset, a standard benchmark in Vision-and-Language Navigation (VLN)~\cite{anderson2018vision} with object manipulation tasks.
In contrast, state-of-the-art methods (e.g.,~\cite{murray2022following,inoue2022prompter}) have obtained approximately 46\%.

Multimodal language understanding methods for object manipulation instructions have been widely investigated~\cite{hatori2018interactively,magassouba2019understanding,magassouba2020multimodal,ishikawa2021target}.
However, in terms of computational complexity, it is often impractical to simply apply them to the above task by introducing the input of a destination candidate.
This is because it is necessary to conduct a significant number of inferences for all combinations of target object candidates and destination candidates.
Assuming 100 target object candidates and 100 destination candidates, 10000 inferences would be needed to determine the maximum likelihood pair.
If a single inference takes $4\times10^{-3}$ seconds, the whole computation will take 40 seconds, which is impractical.

In this paper, we propose Switching Head--Tail Funnel UNITER (SHeFU), which can solve the task by predicting the target objects and destinations individually using a single model.
The computational complexity of SHeFU is $O(M+N)$ rather than $O({M}\times{N})$, where $M$ and $N$ denote the numbers of target object candidates and destination candidates, respectively.
Unlike existing methods, we introduce the Switching Head--Tail mechanism to handle both target object candidates and destination candidates by a single model.
The Switching Head mechanism conditions the model by implicitly sharing the parameters required to predict the target object and the destination.
In contrast, the Switching Tail mechanism enables multi-task learning.
These mechanisms use visual and linguistic information about the destination when predicting the target object, and vice-versa.
In addition, there is no need to prepare separate models because both tasks can be handled by a single model.
The main contributions of this study are summarized as follows:
\begin{itemize}
    \item We propose a new multimodal language comprehension model called SHeFU. Its computational complexity is $O(M+N)$ rather than $O({M}\times{N})$, where $M$ and $N$ denote the numbers of target object candidates and destination candidates, respectively.
    \item We introduce the Switching Head--Tail mechanism, which enables both the target objects and destinations to be predicted individually using a single model.
\end{itemize}

%% file: section2.tex
\section{
    Related Work
    \label{related}
}
\vspace{-1mm}

\subsection{Embodied AI}
\vspace{-1mm}

There has been considerable research in Embodied AI, including benchmarking competitions for DSRs in standardized domestic environments (e.g., RoboCup@Home~\cite{iocchi2015robocup}, World Robot Summit~\cite{okada2019competitions}), which are closely related to the task addressed in this study.
Unlike these competitions, we do not use template-based instruction sentences.

Most existing methods involving natural language understanding focus on navigation (e.g.,~\cite{shah2022lmnav,huang23vlmaps}), object manipulation (e.g.,~\cite{shridhar2022cliport,zhou2022modularity}), or a combination of the two (e.g.,~\cite{khandelwal2022simple,ahn2022can}).
PaLM-SayCan~\cite{ahn2022can} grounds large language models via value functions of pretrained skills, enabling physical robots to execute long-horizon tasks according to abstract natural language instructions.

Representative Embodied AI tasks include VLN and Object Goal Navigation (ObjNav)~\cite{anderson2018evaluation}.
VLN tasks require a robot to execute navigation and/or object manipulation according to natural language instructions in 3D environments.
The ALFRED~\cite{shridhar2020alfred} dataset can be used to train a mapping from natural language instructions and egocentric vision to sequences of robotic actions for household tasks.
For VLN tasks, transformers~\cite{vaswani2017attention} have recently achieved performance improvements (e.g.,~\cite{chen2022think,ishikawa2022moment}).
For example, \cite{ishikawa2022moment} introduces a moment-based adversarial training algorithm to VLN.
ObjNav tasks aim to navigate a robot to a target object specified by a word with reference to the egocentric vision.
Deep reinforcement learning navigation methods, which use semantic and spatial knowledge regarding objects, have been successfully applied to such tasks (e.g.,~\cite{du2021vtnet,fukushima2022object}).

\subsection{Multimodal Language Processing}
\vspace{-1mm}

There have been many studies on multimodal language processing~\cite{mogadala2021trends,uppal2022multimodal,chen2023vlp}.
For example, \cite{uppal2022multimodal} presents a survey of vision-and-language studies, providing a comprehensive overview of the latest trends in various tasks and methods.
The field of multimodal language processing can be divided into sub-fields depending on the combination of modalities.
The main vision-and-language sub-fields include Referring Expression Comprehension (REC) (e.g.,~\cite{chen2020uniter,wang2022ofa}), Referring Expression Segmentation (RES) (e.g.,~\cite{yang2022lavt,wang2022cris}, and Multimodal Language Understanding for Fetching Instructions (MLU-FI).

REC involves grounding a target object to a bounding box, described by a natural language that contains a referring expression.
Public datasets for REC tasks include RefCOCO~\cite{yu2016modeling} and its variants.
The RES task is a derivative of REC in which pixel-wise classification is applied to segment out a referenced region instead of a bounding box.
The MLU-FI task is closely related to the task addressed in this study.
In this task, models predict the bounding box of the target object described by the object manipulation instruction through binary classification of whether the candidate target object is truly the target object or not.
The datasets for the MLU-FI task include PFN-PIC~\cite{hatori2018interactively}, WRS-PV~\cite{ogura2020alleviating}, and WRS-UniALT~\cite{ishikawa2021target}.
Each of these datasets consists of images and instructions about the target objects to be grasped.
The PFN-PIC dataset contains images of approximately 20 different everyday objects, randomly placed in four boxes.
The images were captured from a fixed viewpoint in a real-world environment.

Most existing methods for the MLU-FI task take whole images~\cite{hatori2018interactively,magassouba2019understanding,magassouba2020multimodal} or individual regions acquired by object detection~\cite{ishikawa2021target} as their input.
Our method is closely related to MLU-FI methods, especially Target-dependent UNITER (TDU)~\cite{ishikawa2021target}.
TDU uses the transformer attention mechanism based on the UNITER~\cite{chen2020uniter} framework to model the relationship between objects and instructions.
However, while the MLU-FI task only requires the identification of target objects from object manipulation instructions, the task addressed in this study includes referring expressions regarding the target object and the destination.
Therefore, SHeFU is different from existing methods in that the Switching Head--Tail mechanism enables the individual prediction of both target objects and destinations using a single model.
Note that the Switching Head mechanism is also different from existing methods that perform multi-task learning by switching feature extraction networks (e.g.,~\cite{chen2021pre}).
In our method, it conditions the model by partially zero-filling the input, as described in Section~\ref{switching_image_embedder}.

%% file: section3.tex
\section{Problem Statement
\label{sec:problem}
}
\vspace{-1mm}

\begin{figure*}[t]
    \centering
    \includegraphics[width=\linewidth]{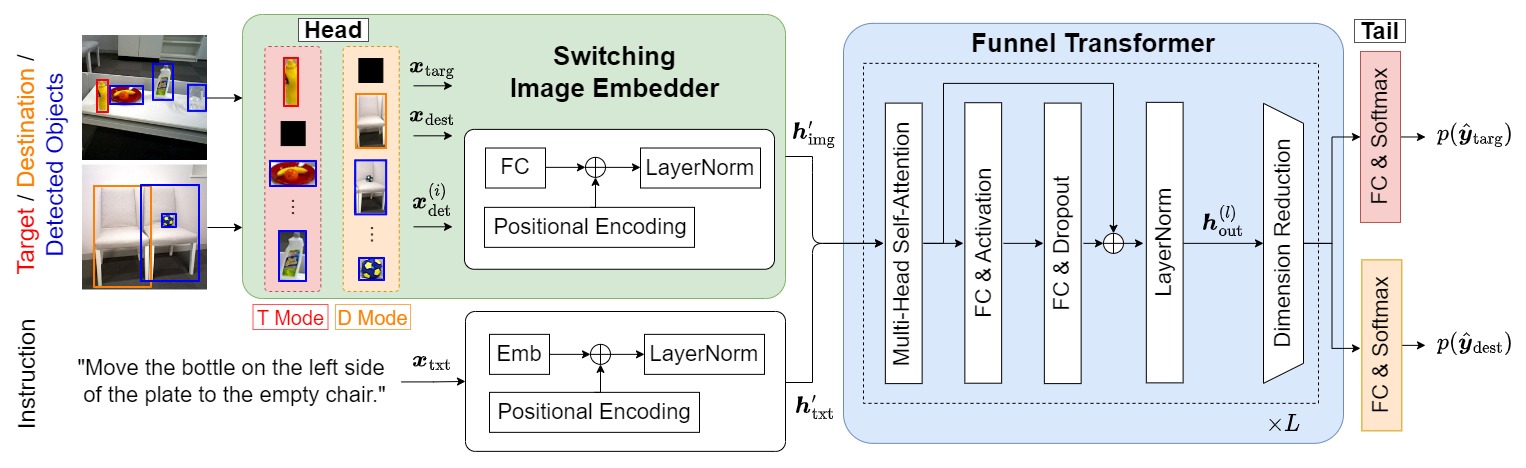}
    \caption{\small Proposed method structure: SHeFU mainly consists of Switching Image Embedder and Funnel Transformer. Here, ``FC,'' ``Emb,'' ``$\oplus$,'' and the merging arrows with rounded corners represent the fully-connected layer, embedding, addition, and concatenation, respectively.}
    \label{fig:model}
    \vspace{-4mm}
\end{figure*}


In this paper, we define the task as the Dual Referring Expression Comprehension with fetch-and-carry (DREC-fc) task.
In this task, given an instruction that contains referring expressions, the DSR identifies both the target object and the destination from multiple candidates of everyday objects or furniture pieces, and subsequently carries the target object to the destination.
Therefore, this task can be divided into two sub-tasks: language comprehension and action execution.


In this task, it is desirable for a DSR to predict whether a target object candidate and a destination candidate match the correct target object and the destination, respectively, and then to deliver the target object to the specified destination.
Fig.~\ref{fig:eye_catch} shows an example of the DREC-fc task.
For the images (a) and (b), consider the instruction ``Move the bottle on the left side of the plate to the empty chair.''
In this case, it is expected that the DSR will identify the red and orange bounding boxes as the target object and the destination, from among the surrounding objects or furniture pieces shown in blue bounding boxes.
The DSR should then execute the corresponding grasping and placing actions.

The terminology used in the remainder of this paper is defined as follows:
\begin{itemize}
    \item \textbf{Target object:} An everyday object to be manipulated.
    \item \textbf{Target object candidate:} An object that the model predicts whether it matches the target object or not.
    \item \textbf{Destination:} A furniture piece on which the target object is to be placed.
    \item \textbf{Destination candidate:} A furniture piece that the model predicts whether it matches the destination or not.
\end{itemize}

The input and output of the task are defined as follows:
\begin{itemize}
    \item \textbf{Input:} An instruction, an image including a target object candidate, and an image including a destination candidate.
    \item \textbf{Output:} A predicted probability $p(\hat{\bm{y}})$ that both the target object candidate and the destination candidate match the respective ground truth.
\end{itemize}

The training of a large-scale model, such as a transformer~\cite{vaswani2017attention}, usually requires a large amount of data.
We use a simulated environment for the efficient collection of data and realize cost-reduction through zero-shot transfer to the real-world environment.
This is because data collection with physical robots is labor-intensive, involving a human experimenter for the placement of objects.

We assume that trajectory generation regarding navigation, object grasping, and placing is based on heuristic methods.
Details are described in Section~\ref{procedure}.

%% file: section4.tex
\section{Proposed Method
\label{method}
}
\vspace{-1mm}


The structure of the proposed method is shown in Fig.~\ref{fig:model}.
Our model mainly consists of two modules: Switching Image Embedder and Funnel Transformer.
Our method is closely related to fetch-and-carry tasks with language instructions, where the robot is requested to bring the target object to the destination according to free-form natural language instructions~\cite{iocchi2015robocup,okada2019competitions}.
Note that template-based sentences are not used in this study.

In this study, we validate the proposed method on the multimodal language comprehension with object manipulation tasks.
The Switching Head--Tail mechanism in our method can also be applied to other tasks, such as natural language comprehension tasks with more than two referring expressions and RES tasks.
The proposed method is applicable to simulated and real-world environments if the input is the same.
Evaluation on a dataset collected in a simulated environment and validation in a real-world environment are described in Sections~\ref{sim_exp} and \ref{real_exp}, respectively.

\subsection{Input}
\vspace{-1mm}

The input to our model is defined as $\bm{x} = \{\bm{x}_\mathrm{targ}, \bm{x}_\mathrm{dest}, \bm{x}_\mathrm{txt}\}$, where $\bm{x}_\mathrm{targ} \in \mathbb{R}^{1024}$, $\bm{x}_\mathrm{dest} \in \mathbb{R}^{1024}$, and $\bm{x}_\mathrm{txt} \in \{0,1\}^{D_v \times D_l}$ denote the target object candidate region, destination candidate region, and instruction, respectively.
$D_v$ and $D_l$ denote the vocabulary size and the maximum number of tokens in the instructions, respectively.

We extract $\bm{x}_\mathrm{targ}$ and $\bm{x}_\mathrm{dest}$ from the output of the fc6 layer of ResNet50~\cite{he2016deep}, which is used as the backbone network of Faster R-CNN~\cite{ren2016faster}.
We use a seven-dimensional vector $[\frac{x_1}{W}, \frac{y_1}{H}, \frac{x_2}{W}, \frac{y_2}{H}, \frac{w}{W}, \frac{h}{H}, \frac{w \cdot h}{W \cdot H}]^T$ as the positional encoding for visual features of the regions.
Here, ($W$, $H$), ($w$, $h$), ($x_1$, $y_1$), and ($x_2$, $y_2$) denote the width and height of the input image, the width and  height of each region, the coordinate of the upper left vertex of each region, and the coordinate of the lower right vertex of each region, respectively.

To obtain $\bm{x}_\mathrm{txt}$, the instruction is tokenized using WordPiece, as in BERT~\cite{devlin2019bert}.
We use the positions of the tokens in the instruction as the positional encoding for text features.
To obtain embedded feature $\bm{h}^{\prime}_\mathrm{txt}$, $\bm{x}_\mathrm{txt}$ are simply multiplied by trainable weights and normalized.


\subsection{
    Switching Image Embedder
    \label{switching_image_embedder}
}
\vspace{-1mm}

This module switches inputs depending on the mode through the Switching Head mechanism and embeds the target object candidate region, destination region, and other detected regions.
The module has two modes, the target mode and the destination mode.
The target object and the destination are predicted in the corresponding modes.
The input to this module consists of $\bm{x}_\mathrm{targ}$ and $\bm{x}_\mathrm{dest}$.

First, the Switching Head mechanism handles $\bm{x}_\mathrm{targ}$ and $\bm{x}_\mathrm{dest}$ as follows:
\begin{align}
    \nonumber
    (\bm{x}_\mathrm{targ}, \bm{x}_\mathrm{dest}) =
    \begin{cases}
        (\bm{x}_\mathrm{targ}, \bm{0}) & \text{if target mode}\\
        (\bm{0}, \bm{x}_\mathrm{dest}) & \text{if destination mode}
    \end{cases}.
\end{align}
In other words, unnecessary input in each mode is filled with zeroes.
This works as a conditioning that switches what is to be predicted.
In addition, we obtain the regions $\{\bm{x}_\mathrm{det}^{(i)} \in \mathbb{R}^{1024} \mid i=1, \ldots, K\}$ of objects and furniture pieces by using Faster R-CNN.
Here, $K$ denotes the maximum number of detected regions in an image.
Note that the image feature extraction and positional encoding are the same for $\bm{x}_\mathrm{targ}$ and $\bm{x}_\mathrm{dest}$.

Next, $\bm{x}_\mathrm{targ}$, $\bm{x}_\mathrm{dest}$, and $\bm{x}_\mathrm{det}^{(i)}$ are input to their respective fully-connected layers, and each output is normalized to obtain $\bm{h}^{\prime}_\mathrm{targ}$, $\bm{h}^{\prime}_\mathrm{dest}$, and ${\bm{h}^{\prime}_\mathrm{det}}^{(i)}$, respectively.
These are concatenated as $\bm{h}^{\prime}_\mathrm{img}=\{\bm{h}^{\prime}_\mathrm{targ}, \bm{h}^{\prime}_\mathrm{dest}, {\bm{h}^{\prime}_\mathrm{det}}^{(1)}, \ldots, {\bm{h}^{\prime}_\mathrm{det}}^{(K)}\}$.

\subsection{Funnel Transformer}
\vspace{-1mm}

This module consists of the $L$-layer Funnel Transformer~\cite{dai2020funnel}.
The input of the first layer is defined as $\bm{h}_\mathrm{in}^{(1)}=\{\bm{h}^{\prime}_\mathrm{img}, \bm{h}^{\prime}_\mathrm{txt}\}$.
The output $\bm{h}_\mathrm{out}^{(l)}$ of the $l$-th layer is computed based on the self-attention mechanism in the transformer~\cite{vaswani2017attention}.
In this process, we use max-pooling to reduce the dimension of $\bm{h}_\mathrm{out}^{(l)}$.
Here, the dimension of the query, key, and value in the $l (>1)$-th layer is computed as $H^{(l)}=\lfloor \frac{H^{(l-1)}}{2} \rfloor$, where $\lfloor \cdot \rfloor$ denotes the floor function.
Similarly, the number of attention heads in the $l (>1)$-th layer is calculated as $A^{(l)}=\lfloor \frac{A^{(l-1)}}{2} \rfloor$.
Note that we apply max-pooling to the query, key, and value for empirical reasons, whereas~\cite{dai2020funnel} only applied it to the query.
Using $\bm{h}_\mathrm{in}^{(l)}=\bm{h}_\mathrm{out}^{(l-1)}$ as the input of the $l$-th layer, $\bm{h}_\mathrm{out}^{(l)}$ is computed in the same way as for the $(l-1)$-th layer.
The output $\bm{h}^{\prime}_\mathrm{out}$ of the Funnel Transformer module is obtained by repeating the above process up to the $L$-th layer.

The module also has two modes.
The Switching Tail mechanism switches the last network according to the mode.
The predicted probability $p(\hat{\bm{y}}_\mathrm{targ})$ with respect to the target object is obtained as follows:
\begin{align}
    p(\hat{\bm{y}}_\mathrm{targ}) = \mathrm{softmax}(f_\mathrm{FC}(\bm{h}^{\prime}_\mathrm{out})),\nonumber
\end{align}
where $f_\mathrm{FC}$ denotes a fully-connected layer.
Similarly, the predicted probability $p(\hat{\bm{y}}_\mathrm{dest})$ with respect to the destination is obtained using a different fully-connected layer.
The final output is $p(\hat{\bm{y}}_\mathrm{targ})$ in the target mode and $p(\hat{\bm{y}}_\mathrm{dest})$ in the destination mode.
We obtain the predicted label $\hat{y}_\mathrm{targ}$ or $\hat{y}_\mathrm{dest}$ in each mode by binarizing the predicted probability using a threshold of 0.5.
Note that the Switching Tail mechanism predicts the target object and the destination individually.
Therefore, the predicted label $\hat{y}$ is obtained as follows:
\begin{align}
    \hat{y} = \hat{y}_\mathrm{targ} \cap \hat{y}_\mathrm{dest}.
    \label{eq:y_hat}
\end{align}

\subsection{Loss Function}
\vspace{-1mm}

We use the loss function $\mathcal{L}$ as follows:
\begin{align}
    \mathcal{L} = \lambda_\mathrm{targ}\mathcal{L}_\mathrm{CE}(\bm{y}_\mathrm{targ}, p(\hat{\bm{y}}_\mathrm{targ})) + \lambda_\mathrm{dest}\mathcal{L}_\mathrm{CE}(\bm{y}_\mathrm{dest}, p(\hat{\bm{y}}_\mathrm{dest})),\nonumber
\end{align}
where $\mathcal{L}_\mathrm{CE}(\cdot, \cdot)$ and $\lambda_\cdot$ denote the cross-entropy loss and the task weights in each mode, respectively.
Here, $y_\cdot$ denotes a Boolean value indicating whether the candidate target object or destination matches the respective ground truth.
Multi-task learning is performed by setting $\lambda_\mathrm{dest}=0$ in the target mode and $\lambda_\mathrm{targ}=0$ in the destination mode.

%% file: section5.tex
\section{
    Simulation Experiments
    \label{sim_exp}
}
\vspace{-1mm}

\subsection{
    Dataset
    \label{dataset}
}
\vspace{-1mm}

In this study, we built the novel ALFRED-fc dataset for the DREC-fc task based on the ALFRED dataset~\cite{shridhar2020alfred}.
This is because, to the best of our knowledge, there is no standard dataset for the DREC-fc task.
ALFRED is a standard dataset for VLN with object manipulation tasks, which enables the training of a mapping from natural language instructions and egocentric vision to sequences of robotic actions for household tasks.
The dataset includes multiple sequential sub-goals for the robot’s behavior and an instruction for each sub-goal.
The instructions in the ALFRED dataset were given by at least three annotators using Amazon Mechanical Turk.
The annotators were asked to give instructions to guide the robot in achieving the sub-goal.

However, the original ALFRED dataset cannot be used directly because most of the camera images of the robot carrying an object contain the target object floating in the air, which is unrealistic.
For this reason, we collected camera image-pairs captured just before an object was grasped and just after it was placed.
They were collected only in the episodes where their sub-goals belong to the ``Pick \& Place'' category.
Here, Pick \& Place is the sub-goal whereby the robot grasps a specific object and places it at a designated location.
One limitation of the ALFRED dataset is that it is impossible to capture an image that contains the destination without the corresponding target object.
This is because the image from which the floating target object is removed always contains the target object placed at the destination.
Therefore, the target object region was masked in the image.

The ALFRED dataset contains ground truth regions with respect to the target object and destination in each image.
However, the regions of other objects and furniture pieces are not given.
Hence, we extracted their regions using Faster R-CNN~\cite{ren2016faster}.
Note that the number of positive and negative samples was balanced.
We created positive samples from detected regions for which the IoU with the ground truth was greater than or equal to 0.7.
Negative samples were created by randomly employing one of the three methods as follows:
\begin{enumerate}
    \setlength{\parskip}{0.5mm} 
    \setlength{\itemsep}{0.2mm} 
    \item Regions with $\mathrm{IoU} \leq 0.3$ were selected as target object candidates.
    \item Instructions were replaced with those of randomly selected samples. This method was employed to prevent trivial solutions in cases where there were several misdetected regions in the images.
    \item Both of the above.
\end{enumerate}

The ALFRED-fc dataset consists of 1099 image-pairs containing a target object and a destination, respectively, and 3452 instructions written in English with a vocabulary size of 646 words, a total of 29113 words, and an average sentence length of 8.4 words.
The ALFRED-fc dataset includes 4420, 642, and 686 samples in the training, validation, and test sets, respectively.
Note that a sample consists of an instruction, an image with respect to a target object candidate, and an image with respect to a destination candidate.
We built the training and validation sets based on the training set of the ALFRED dataset.
The test set was constructed based on both the seen and unseen splits of the validation set of the ALFRED dataset.
We used the training set to train the model and the validation set to tune the hyperparameters.
We evaluated our model on the test set.

\subsection{Parameter Settings}
\vspace{-1mm}


In the Funnel Transformer module, we set $L=2$, $H^{(1)}=(K+D_l+1)\times768$, and $A^{(1)}=12$, where $K$ and $D_l$ denote the maximum number of objects or furniture pieces in the image and the maximum number of tokens in the instructions, respectively.
The task weights were set to $\lambda_\mathrm{targ}=1.0$ and $\lambda_\mathrm{dest}=1.0$.
We adopted the Adam optimizer ($\beta_1=0.9$, $\beta_2=0.999$) for training with learning rate $8\times10^{-5}$, batch size 8, and dropout probability 0.1.

Our model had approximately 33 M trainable parameters.
We trained our model on a GeForce RTX 3090 with 24 GB of GPU memory and an Intel Core i9-10900KF with 64 GB of RAM.
It took approximately 20 minutes for training.
The inference time was approximately $4\times10^{-3}$ seconds for one sample.
Throughout a total of 20000 training steps, we measured the accuracy on the validation set every 2000 training steps.
The final performance was given by the test set accuracy when the validation set accuracy was maximized.

\subsection{Quantitative Results}
\vspace{-1mm}

\input{tab/quantitative_language}

Table~\ref{tab:quantitative_language} presents the quantitative results.
The table compares the accuracy of each method on the ALFRED-fc dataset.
These results are the mean and standard deviation of the accuracy over five experimental runs.

We used an extended version of TDU~\cite{ishikawa2021target} as the baseline method.
This is because it has been successfully applied to the MLU-FI task, which is closely related to the DREC-fc task.
The baseline method includes both a target object candidate and a destination candidate as input.

We employed accuracy as the evaluation metric.
This is because accuracy is a standard metric when the number of positive and negative samples is balanced, and the dataset used in the experiments satisfied this condition.
Unlike the baseline method, the proposed method predicts the target object and the destination individually.
To compare the two methods in a fair manner, we define the true label as $y = y_\mathrm{targ} \cap y_\mathrm{dest}$.

According to Table~\ref{tab:quantitative_language}, the proposed method (iv) achieved an accuracy of 83.1\%, compared with 79.4\% for the baseline method (i).
Thus, our method outperformed the baseline method by 3.7 points.
This difference in performance is statistically significant ($p$-value $< 0.01$).

\subsection{Qualitative Results}
\vspace{-1mm}

\begin{figure}
    \centering
    \begin{tabular}{c}
        \small (a)
        \begin{minipage}{0.45\hsize}
            \centering
            \includegraphics[width=\linewidth]{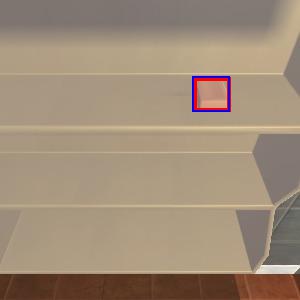}
        \end{minipage}
        \begin{minipage}{0.45\hsize}
            \centering
            \includegraphics[width=\linewidth]{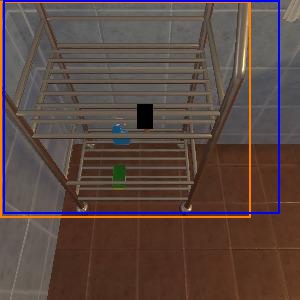}
        \end{minipage}\\
        \small Instruction: ``Move the soap from the shelves to the metal rack.''\\
        \small (b)
        \begin{minipage}{0.45\hsize}
            \centering
            \includegraphics[width=\linewidth]{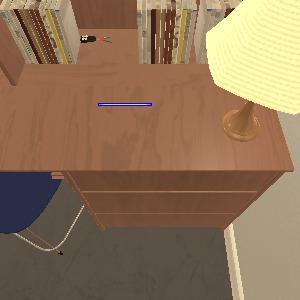}
        \end{minipage}
        \begin{minipage}{0.45\hsize}
            \centering
            \includegraphics[width=\linewidth]{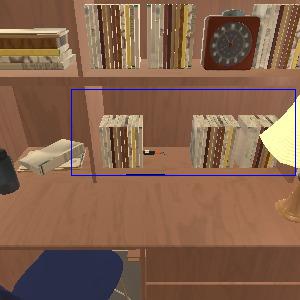}
        \end{minipage}\\
        \small Instruction: ``Put a salt shaker into a kitchen drawer.''
    \end{tabular}
    \caption{\small Qualitative results on the ALFRED-fc dataset. Panels (a) and (b) show TP and TN samples, respectively. From left to right: target object candidate and destination candidate. Red, orange, and blue bounding boxes represent the ground truth region of the target object, the ground truth region of the destination, and the target object or destination candidates, respectively.}
    \label{fig:qualitative_alfred-fc}
    \vspace{-4mm}
\end{figure}

Fig.~\ref{fig:qualitative_alfred-fc} shows the qualitative results on the ALFRED-fc dataset.
Case (a) in Fig.~\ref{fig:qualitative_alfred-fc} shows a True Positive (TP) sample.
The target object and the destination were the soap on the shelves and the metal rack, respectively.
Both the target object and destination candidates matched their respective ground truth regions.
For this sample, while the baseline model incorrectly predicted that $\hat{y}=0$, the proposed model correctly predicted that $(\hat{y}_\mathrm{targ}, \hat{y}_\mathrm{dest})=(1, 1)$.

Similarly, Case (b) in Fig.~\ref{fig:qualitative_alfred-fc} shows a True Negative (TN) sample.
The target object and the destination were the salt shaker and the kitchen drawer, respectively.
Neither the target object candidate nor the destination candidate matched their respective ground truth regions.
For this sample, the proposed model correctly predicted that $(\hat{y}_\mathrm{targ}, \hat{y}_\mathrm{dest})=(0, 0)$, whereas the baseline model mistakenly predicted that $\hat{y}=1$.
Our model correctly predicted that both the pen on the desk and the bookshelf were irrelevant to the instruction.

\subsection{Ablation Study}
\vspace{-1mm}

For an ablation study, we set the two conditions as follows:
\begin{enumerate}
    \setlength{\parskip}{0.5mm} 
    \setlength{\itemsep}{0.2mm} 
    \renewcommand{\labelenumi}{(\roman{enumi})}
    \setcounter{enumi}{1}
    \item W/o Switching Head: To investigate the contribution of the Switching Head mechanism to the accuracy improvement, we set $\bm{x}_\mathrm{targ}$ and $\bm{x}_\mathrm{dest}$ as follows:
    \begin{align}
        \nonumber
        (\bm{x}_\mathrm{targ}, \bm{x}_\mathrm{dest}) =
        \begin{cases}
            (\bm{x}_\mathrm{targ}, \bm{x}_\mathrm{targ}) & \text{if target mode} \\
            (\bm{x}_\mathrm{dest}, \bm{x}_\mathrm{dest}) & \text{if destination mode}
        \end{cases}.
    \end{align}
    \item W/o Switching Tail: To investigate the contribution of multi-task learning in the Switching Tail mechanism to the accuracy improvement, single-task learning on the target objects and destinations was conducted by a single model.
\end{enumerate}

Table~\ref{tab:quantitative_language} shows that the accuracy under conditions (ii) and (iii) were lower than when using the proposed method (iv) by 4.7 points and 6.2 points, respectively.
These differences in performance are both statistically significant ($p$-value $< 0.01$).
This indicates that both the Switching Head and Tail mechanisms are effective, with the latter having the greatest influence on performance.

\subsection{Error Analysis and Discussion}
\vspace{-1mm}


On the test set, there were 453, 662, 24, and 233 samples classified as TP, TN, FP, and False Negative, respectively.
Thus, a total of 257 samples were incorrectly predicted by the proposed method.

\input{tab/error}

We manually analyzed a total of 100 failure samples.
Table~\ref{tab:error} lists the categories of the failure cases that were incorrectly predicted by the proposed method on the ALFRED-fc dataset.
Here, the proposed method obtained a predicted label $\hat{y}$ based on the target and destination modes according to \eqref{eq:y_hat}.
Therefore, we classified the categories of failure cases in each mode individually.
The causes of failure can be roughly divided into five types: serious comprehension errors (SC), similar object or furniture (SOF), tiny region (TR), insufficient visual information (IVI), and others.
The SC category refers to cases where there was little feature similarity between the candidate and ground truth regions.
The SOF category refers to cases where the target object and the target object candidate, or the destination and the destination candidate, were similar.
The TR category refers to cases where the cause of failure was that the candidate region was too small to be accurately identified.
The IVI category refers to cases where the candidate region did not fully enclose the object or piece of furniture, making it difficult to detect the visual features.
``Others'' category refers to cases where the causes of failure did not fit in any of the above categories (e.g., annotations were incorrectly given, instructions contained incomplete information).

Table~\ref{tab:error} indicates that the main bottleneck was SC in both the target and destination modes.
A possible solution for overcoming these issues is to introduce CLIP~\cite{radford2021learning}, a vision and language model pretrained on large-scale datasets.

%% file: tab/quantitative_language.tex
\begin{table}
    \normalsize
    \caption{\small Language comprehension accuracy on the \\ALFRED-fc dataset and the real-world environment}
    \label{tab:quantitative_language}
    \centering
    \scalebox{.95}{
        \begin{tabular}{m{0.5em}lcc}
            \hline
            [\%] & \;Method & ALFRED-fc & Real\\
            \hline\hline
            (i) & \;Baseline (extended TDU~\cite{ishikawa2021target}) & $79.4 \pm 2.76$ & $52.0$\\
            (ii) & \;Ours (W/o Switching Head) & $78.4 \pm 2.05$ & -\\
            (iii) & \;Ours (W/o Switching Tail) & $76.9 \pm 2.91$ & -\\
            (iv) & \;Ours (SHeFU) & $\mathbf{83.1 \pm 2.00}$ & $\mathbf{55.9}$\\
            \hline
        \end{tabular}
    }
    \vspace{-4mm}
\end{table}

%% file: tab/error.tex
\begin{table}[t]
    \caption{\small Categories of failure cases on the ALFRED-fc dataset}
    \label{tab:error}
    \centering
    \begin{tabular}{ccc}
        \hline
        \begin{tabular}{c}
            Error ID
        \end{tabular} &
        \begin{tabular}{c}
            Target Mode
        \end{tabular} &
        \begin{tabular}{c}
            Destination Mode
        \end{tabular} \\
        \hline\hline
        SC &
        34 & 25 \\
        SOF &
        8 & 7 \\
        TR &
        7 & 0 \\
        IVI &
        0 & 15 \\
        Others &
        1 & 3 \\
        \hline\hline
        Total &
        50 & 50 \\
        \hline
    \end{tabular}
    \vspace{-5mm}
\end{table}

%% file: section6.tex
\section{
    Physical Experiments
    \label{real_exp}
}
\vspace{-1mm}

\subsection{Environment}
\vspace{-1mm}

\begin{figure}[t]
    \centering
    \includegraphics[height=3.4cm]{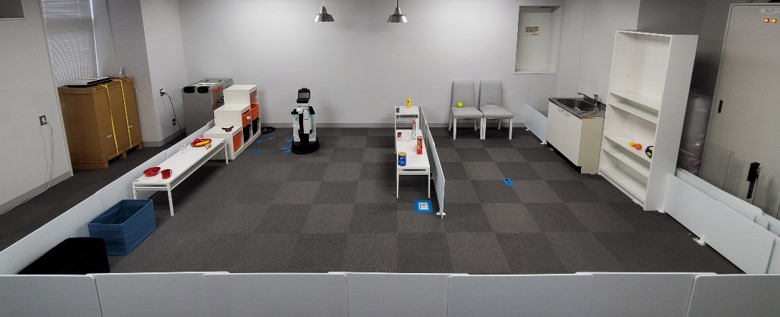}
    \caption{\small Experimental environment. The specification of the environment is standardized and specified in~\cite{wrs2020}.}
    \label{fig:environment}
    \vspace{-1mm}
\end{figure}

Fig.~\ref{fig:environment} shows the environment used in the physical experiments.
This environment was based on the standardized environment of the World Robot Summit 2020 Partner Robot Challenge/Real Space (WRS2020RS)~\cite{wrs2020}, an international robotics competition that benchmarked tidying tasks in the domestic environment.
The size of the environment was $6.0 \times 4.0$ $\mathrm{m}^2$.
There were six types of furniture pieces specified in the WRS2020RS rulebook.
The arrangement of the furniture pieces was as shown in Fig.~\ref{fig:environment}.
Note that there were two storage boxes of different colors and two identical long tables and chairs.
Thus, there were a total of nine pieces of furniture in the environment.
In this experiment, a piece of furniture was randomly selected as the destination.

\subsection{DSR and Objects}
\vspace{-1mm}

\begin{figure}[t]
    \centering
        \begin{tabular}{c}
            \begin{minipage}{0.4\columnwidth}
                \centering
                \includegraphics[height=4.9cm]{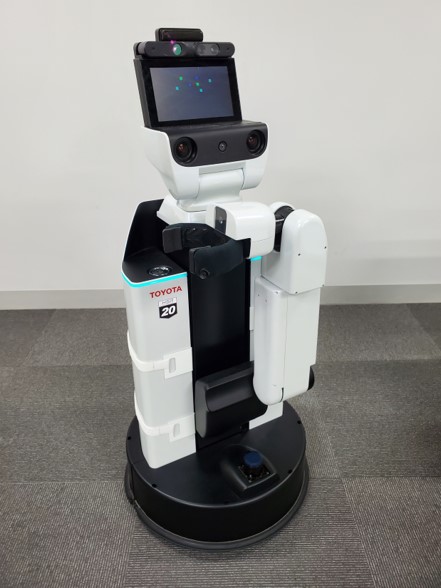}\\
                \small (a)
            \end{minipage}
            \begin{minipage}{0.6\columnwidth}
                \centering
                \includegraphics[height=4.9cm]{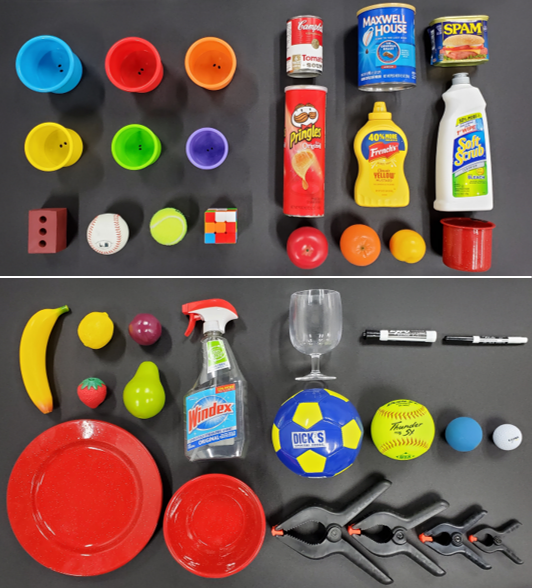}\\
                \small (b)
            \end{minipage}
        \end{tabular}
    \caption{\small (a) DSR platform~\cite{yamamoto2019development} and (b) objects~\cite{calli2015benchmarking} used in the physical experiments.}
    \label{fig:hsr_objects}
    \vspace{-4mm}
\end{figure}

We used the Human Support Robot (HSR)~\cite{yamamoto2019development} developed by the Toyota Motor Corporation, shown in Fig.~\ref{fig:hsr_objects} (a).
Fig.~\ref{fig:hsr_objects} (b) shows the objects used in the physical experiments.
These objects are included in the YCB objects~\cite{calli2015benchmarking}.
The upper group of 20 objects and the lower group of 19 objects were utilized as target objects and background objects, respectively.
We selected target objects that can be grasped by the HSR end-effector from the categories of ``Food,'' ``Kitchen,'' ``Shape,'' and ``Task'' in~\cite{calli2015benchmarking}.
The background objects were randomly chosen from the remaining objects.

\begin{figure*}[t]
    \centering
    \begin{tabular}{c}
        \small (a)
        \begin{minipage}{0.212\hsize}
            \centering
            \includegraphics[width=\linewidth]{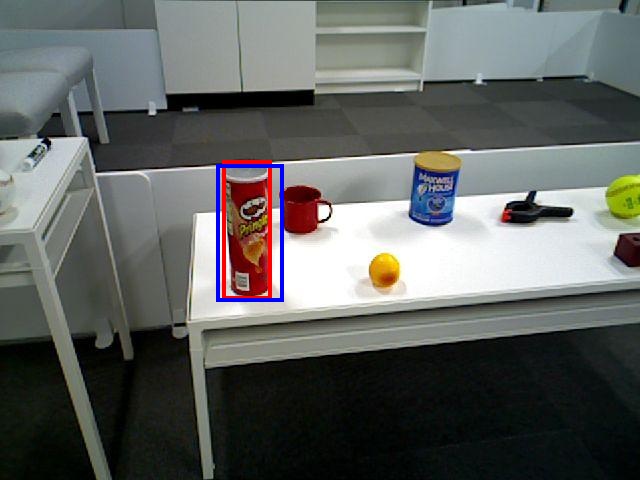}
        \end{minipage}
        \begin{minipage}{0.212\hsize}
            \centering
            \includegraphics[width=\linewidth]{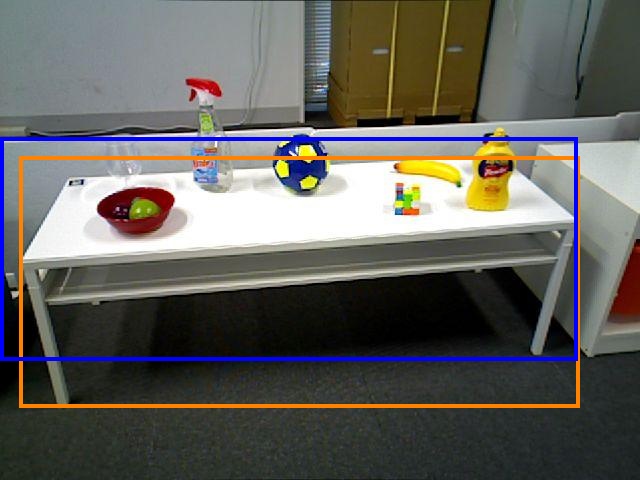}
        \end{minipage}
        \begin{minipage}{0.212\hsize}
            \centering
            \includegraphics[width=\linewidth]{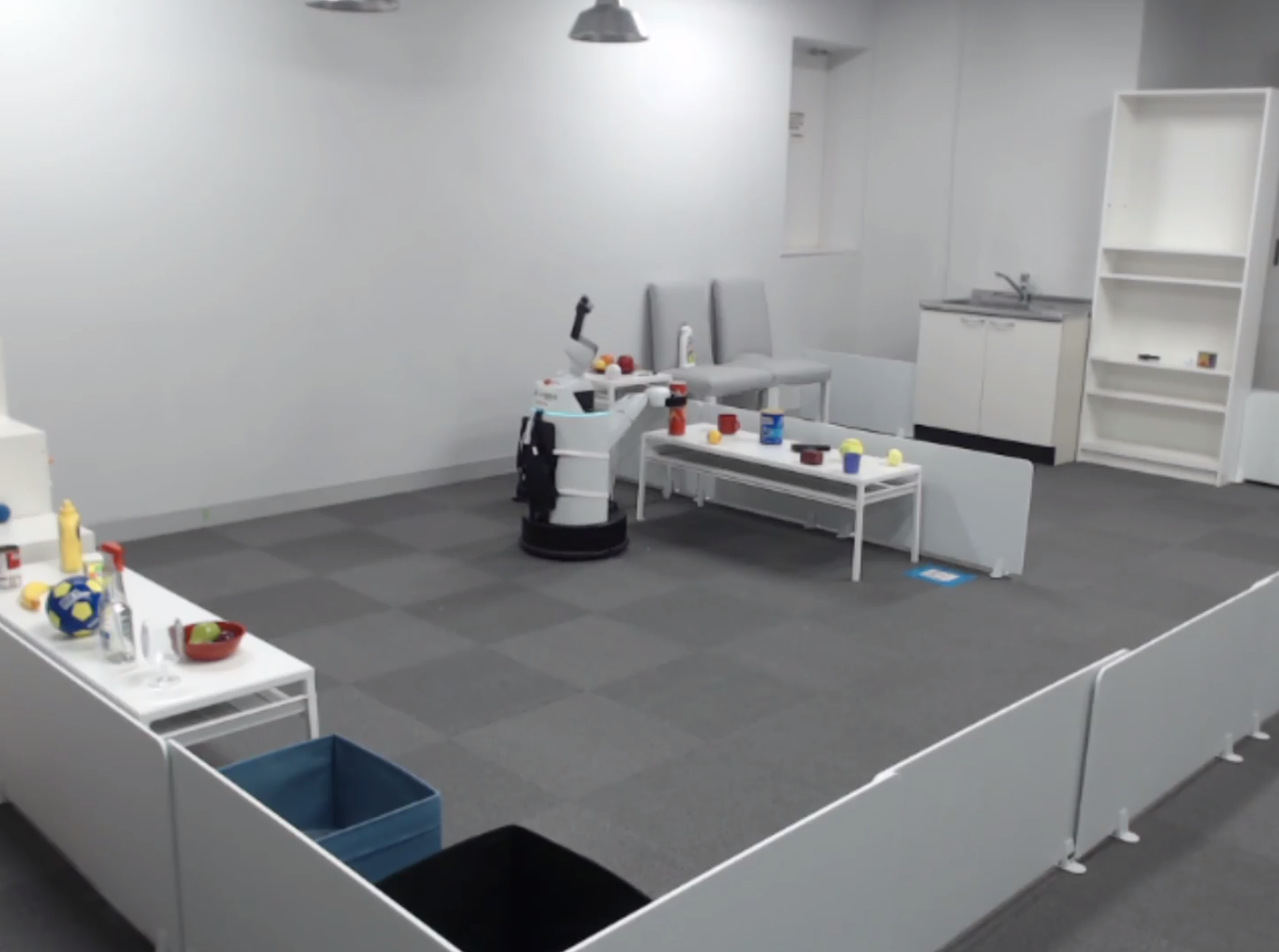}
        \end{minipage}
        \begin{minipage}{0.212\hsize}
            \centering
            \includegraphics[width=\linewidth]{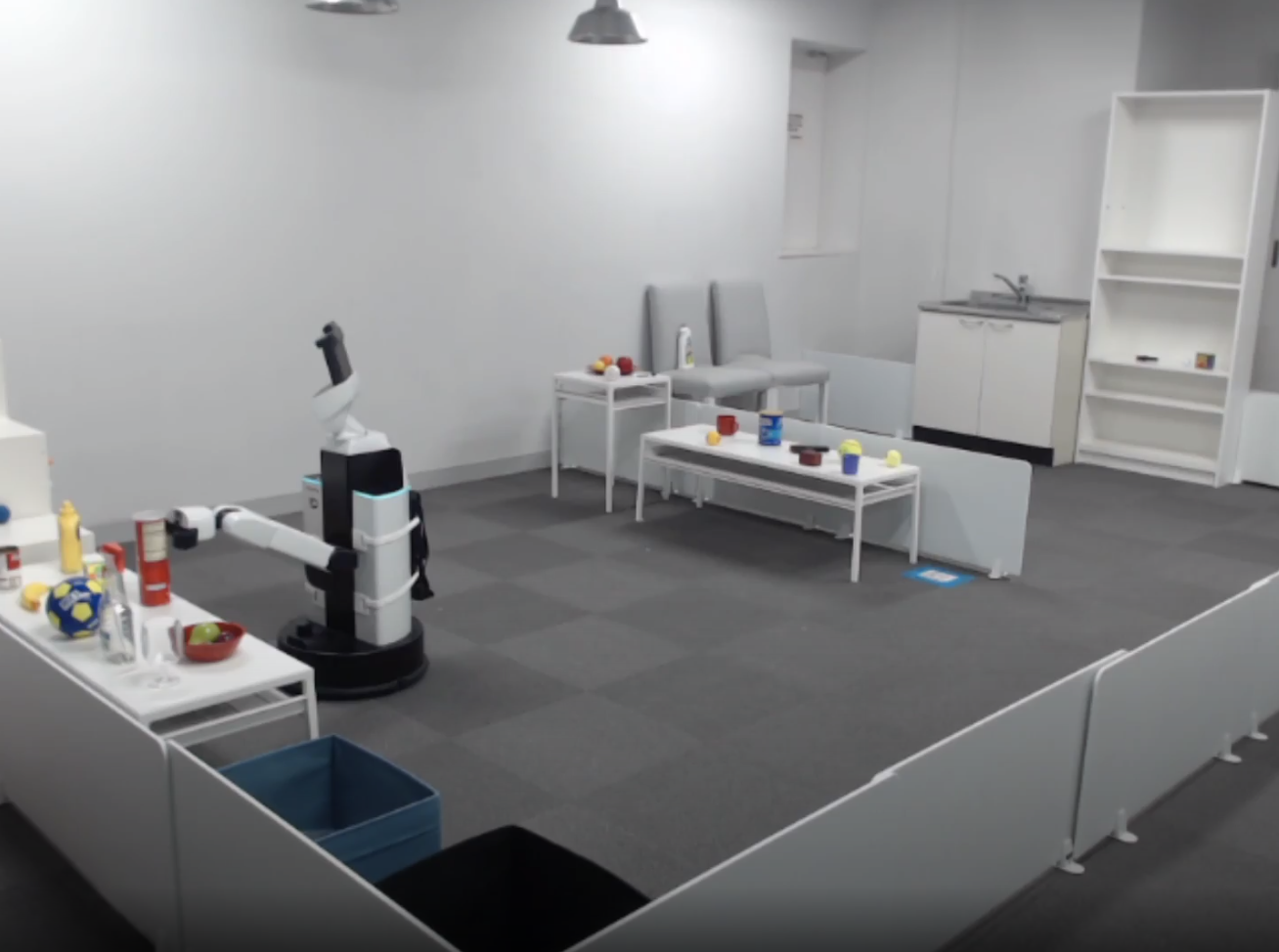}
        \end{minipage}\\
        \small Instruction: ``Put the red chips can on the white table with the soccer ball on it.''\\
        \small (b)
        \begin{minipage}{0.212\hsize}
            \centering
            \includegraphics[width=\linewidth]{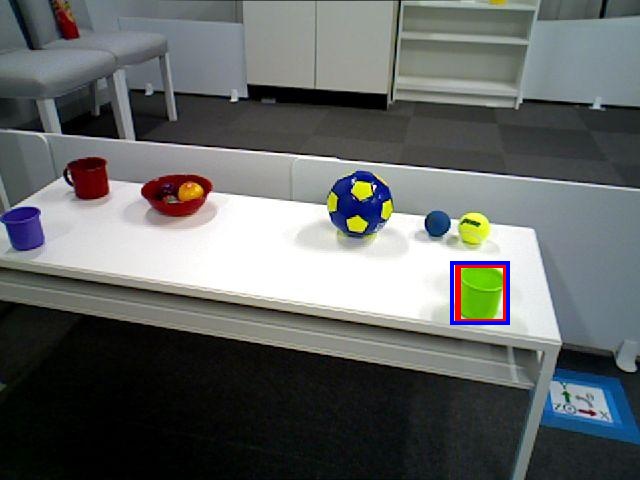}
        \end{minipage}
        \begin{minipage}{0.212\hsize}
            \centering
            \includegraphics[width=\linewidth]{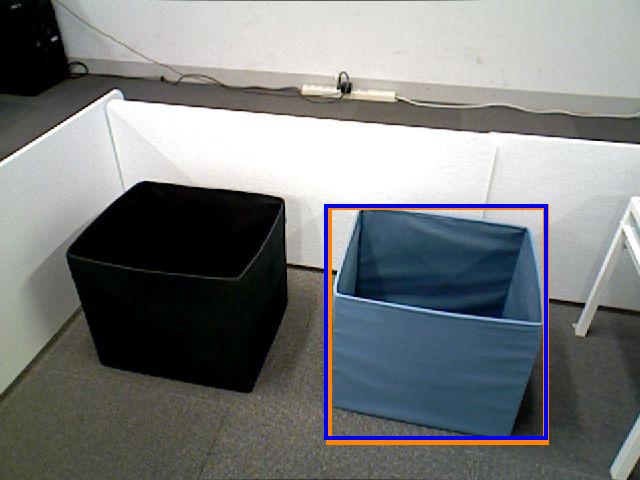}
        \end{minipage}
        \begin{minipage}{0.212\hsize}
            \centering
            \includegraphics[width=\linewidth]{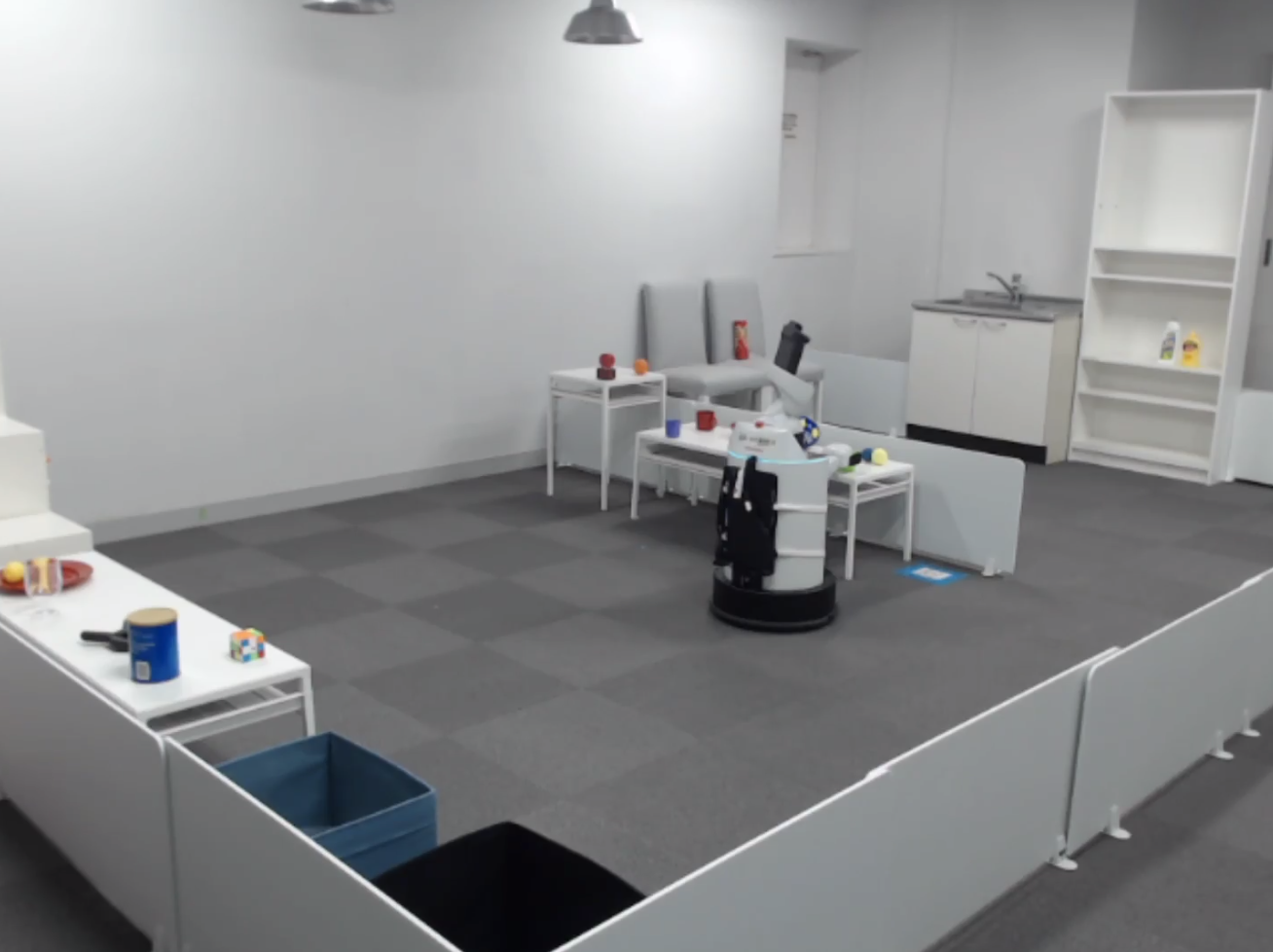}
        \end{minipage}
        \begin{minipage}{0.212\hsize}
            \centering
            \includegraphics[width=\linewidth]{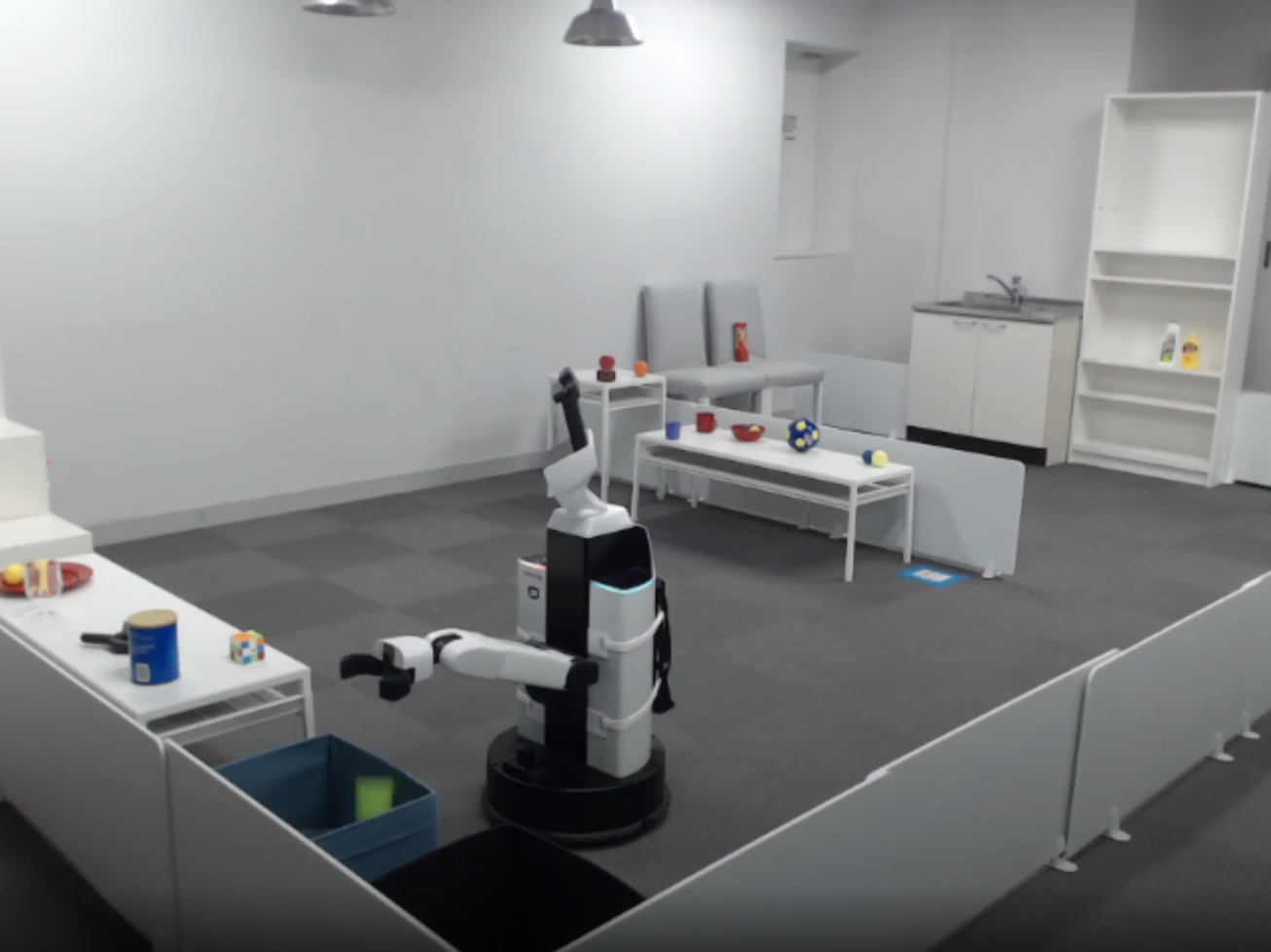}
        \end{minipage}\\
        \small Instruction: ``Place the green cup in the blue bin.''
    \end{tabular}
    \vspace{-0.5mm}
    \caption{\small Qualitative results of the physical experiments. Rows (a) and (b) show successful cases. From left to right: target object candidate, destination candidate, scene of object grasping, and scene of object placing. Red, orange, and blue bounding boxes represent the ground truth region of the target object, the ground truth region of the destination, and the target object or destination candidates, respectively.}
    \label{fig:tp_physical}
    \vspace{-5mm}
\end{figure*}

\subsection{
    Procedure
    \label{procedure}
}
\vspace{-1mm}

In the following, the environment utilized in the physical experiments is explained.
In these experiments, 12 different object placement patterns were created.
The objects were placed in random positions in each object placement pattern.
Note that we assume that all objects in the experiments were placed on furniture pieces.
In each trial, the target object and the destination were randomly selected from the upper group of objects in Fig.~\ref{fig:hsr_objects} (b) and the furniture pieces in Fig.~\ref{fig:environment}, respectively.
Instructions such as ``Pick up the apple and put it down on the right-hand chair'' were then given to the DSR.
A total of 418 instructions were given in English.

The behavior of the DSR was designed as follows.
The DSR first collected images of the environment by moving through 16 predefined waypoints, starting from the initial position.
Path planning and navigation were performed based on standard methods using the map provided in advance.
Each waypoint was defined so that the DSR could face each piece of furniture and capture the object and the furniture pieces from multiple viewpoints.
The DSR captured images using an Asus Xtion Pro camera mounted on its head.
These images were used as input for the proposed model.
We performed zero-shot transfer using our model trained on the ALFRED-fc dataset.
For the grasping action, the grasping point was determined based on the depth image and the bounding box of the target object.
Specifically, a point cloud was obtained from the bounding box of the depth image, and this was transformed to the camera coordinate system by multiplying the intrinsic camera parameters.
The median position in each coordinate axis was determined as the grasping point.
Note that the DSR only attempted to place objects if it had succeeded in the grasping task.
The placing action was based on a rule-based method using the waypoints utilized to capture images of the furniture pieces.

\vspace{-0.7mm}
\subsection{Quantitative Results}
\vspace{-1mm}

Tables~\ref{tab:quantitative_language} and \ref{tab:quantitative_action} present the quantitative results of the physical experiments.
We use the language comprehension accuracy and task success rate (SR) as metrics.
The SR is defined as $\mathrm{SR} = \frac{N_\mathrm{success}}{N_\mathrm{attempts}}$, where $N_\mathrm{attempts}$ and $N_\mathrm{success}$ denote the number of attempts and successes, respectively.

Table~\ref{tab:quantitative_language} compares the language comprehension accuracy.
To evaluate the language comprehension performance, negative samples were created as explained in Section~\ref{dataset}.
Table~\ref{tab:quantitative_language} shows that the proposed method (iv) achieved an accuracy of 55.9\%, compared with 52.0\% for the the baseline method (i).
Thus, our method outperformed the baseline method by 3.9 points.

\input{tab/quantitative_action}

Table~\ref{tab:quantitative_action} presents the SR of the object grasping and placing tasks using the physical robot.
Note that we made the DSR execute the grasping and placing actions only when the predictions were classified as a TP in the language comprehension task.
Trajectory generation regarding object grasping and placing did not use a learning-based method because this is beyond the scope of this study.
Nonetheless, Table~\ref{tab:quantitative_action} indicates that it is possible to integrate language comprehension and action execution in the physical robot.

\vspace{-0.731mm}
\subsection{Qualitative Results and Discussion}
\vspace{-1mm}

Fig.~\ref{fig:tp_physical} shows the qualitative results of the physical experiments.
In Fig.~\ref{fig:tp_physical} (a), the target object and destination were the red chips can and the white table with the soccer ball, respectively.
The equation $(y_\mathrm{targ}, y_\mathrm{dest})=(1, 1)$ holds because both the target object candidate and the destination candidate matched their respective ground truth regions.
For this sample, the proposed model correctly predicted that $(\hat{y}_\mathrm{targ}, \hat{y}_\mathrm{dest})=(1, 1)$.
Subsequently, the DSR accurately grasped the chips can and successfully placed it on the table.

Similarly, in Fig.~\ref{fig:tp_physical} (b), the target object and destination were the green cup and the blue bin, respectively.
The equation $(y_\mathrm{targ}, y_\mathrm{dest})=(1, 1)$ holds because both the target object candidate and the destination candidate matched their respective ground truth regions.
The proposed model correctly predicted that $(\hat{y}_\mathrm{targ}, \hat{y}_\mathrm{dest})=(1, 1)$ for this sample.
Thereafter, the DSR accurately grasped the cup and successfully placed it in the bin.

In a typical scene, 73 target object candidates and 89 destination candidates were detected on average.
Because a single inference took approximately $4\times10^{-3}$ seconds, the computational times for the baseline and proposed methods are considered to be 26 seconds (6497 inferences) and 0.6 seconds (162 inferences), respectively.

%% file: tab/quantitative_action.tex
\begin{table}[t]
    \caption{\small Task success rates in the physical experiments}
    \label{tab:quantitative_action}
    \centering
    \vspace{-1mm}
    \begin{tabular}{lcc}
        \hline
        Task & $N_\mathrm{success} / N_\mathrm{attempts}$ & SR [\%]\\
        \hline\hline
        Grasping & $60$ / $63$ & $95$\\
        Placing & $56$ / $60$ & $93$\\
        \hline
    \end{tabular}
    \vspace{-5.45mm}
\end{table}

%% file: section7.tex
\section{Conclusions}
\vspace{-1mm}
In this study, we focused on the DREC-fc task, in which a DSR identifies both the target object and the destination from multiple candidates according to free-form natural language instructions, and subsequently carries the target object to the destination.
Our contributions are as follows:
\begin{itemize}
    \item We proposed a new multimodal language comprehension model, SHeFU. Its computational complexity is $O(M+N)$ rather than $O({M}\times{N})$, where $M$ and $N$ denote the numbers of target object candidates and destination candidates, respectively.
    \item We introduced the Switching Head--Tail mechanism, which enables both target objects and destinations to be predicted individually using a single model.
    \item SHeFU outperformed the baseline method in terms of language comprehension accuracy on ALFRED-fc, a dataset for the DREC-fc task.
    \item In physical experiments, SHeFU also outperformed the baseline method in terms of language comprehension accuracy. Furthermore, the results indicate that language comprehension and action execution can be integrated in a physical robot.
\end{itemize}
